\documentclass[conference]{IEEEtran}
\usepackage{microtype}
\usepackage{amsmath,amssymb}
\usepackage{booktabs}
\usepackage{graphicx}
\usepackage{xcolor}
\usepackage{url}
\usepackage{enumitem}
\usepackage{caption}
\usepackage{balance}
\usepackage[hidelinks]{hyperref}
\setlist{nosep,leftmargin=*}
\newcommand{\best}[1]{{\fontencoding{OT1}\fontfamily{cmr}\fontseries{bx}\fontsize{10}{10}\selectfont #1}}
\newcommand{\strong}[1]{\best{#1}}
\title{Permutation-Robust Decision Modeling with Candidate-Independent Block-Causal Attention}
\author{Guy Amit\\
Independent Researcher\\
\texttt{guy5@post.bgu.ac.il}}
\date{}

\begin{document}
\maketitle

\begin{abstract}
Decision models often score a variable-sized set of candidate actions encoded in a single sequence. This setting is increasingly relevant for System 1 components inside generative systems, where candidates may be proposed or ordered differently across runs. Standard causal cross-encoding is expressive, but it can make a candidate's score depend on serialization order rather than on the underlying decision problem. We introduce candidate-independent block-causal attention, which preserves causal computation within the shared context and each candidate while blocking cross-candidate information flow and resetting candidate positions. We compare this architecture with standard causal attention and complementary invariant baselines across Gemma 3 1B, Qwen3 1.7B, and Qwen3 4B backbones. Candidate-independent attention consistently reduces permutation sensitivity while retaining competitive decision quality; ablations indicate that candidate isolation is the primary source of the effect, with position resetting completing the intended symmetry. A larger Qwen3-4B study further examines the behavior of the proposed architecture with substantially more training data. Code is available at the \href{https://github.com/guyAmit/ci-decision-models}{\textcolor{blue}{project repository}}, and the \href{https://huggingface.co/Guy-Amit/qwen3-4b-ci-decision-4096-poc}{\textcolor{blue}{Qwen3-4B model artifact}} is available on Hugging Face.
\end{abstract}

\section{Introduction}
Many decision systems receive a state or instruction followed by a variable number of candidate actions and must select one. A convenient implementation concatenates the prompt and candidates and applies a pretrained causal transformer. This cross-encoder can compare candidates through shared hidden states, but serialization order becomes a latent feature: later candidates can condition on earlier candidates, and rotary positions distinguish candidates even when the underlying decision problem is a set.

This issue is especially important when a decision model serves as a fast System 1 component inside a larger generative model. Unlike a conventional classifier with a fixed label ordering, a generative system may propose, retrieve, or assemble candidates at runtime. The resulting order is runtime-dependent and may differ across decoding runs or upstream implementations, even when the underlying candidate set is unchanged. Typed-decision benchmarks such as Open-Jev~\cite{openjev} capture this setting: the model evaluates a displayed set of possible actions rather than predicting from a fixed global label vocabulary. An order-sensitive decision model can therefore produce different scores or selections for the same set of available actions.

We evaluate a simple structural alternative. The shared prefix is processed causally. Candidate block $i$ can attend to the prefix and to its own preceding tokens, but cannot attend to any other candidate. Candidate blocks restart at the same logical positions. Candidate logits still compete through a final softmax, so the model retains a trainable decision head without allowing candidate-specific information flow. We call this model candidate-independent (CI).

Our contributions are: (i) a matched comparison of causal and CI decision heads across three decoder backbones; (ii) group-aware paired evaluation of quality and permutation robustness; (iii) a Gemma 3 1B mechanism ablation separating masking, positions, and permutation augmentation; and (iv) an exploratory comparison to a frozen dual encoder and a larger-data CI validation study.

\section{Background}

\begin{figure*}[t]
\centering
\includegraphics[width=.68\textwidth]{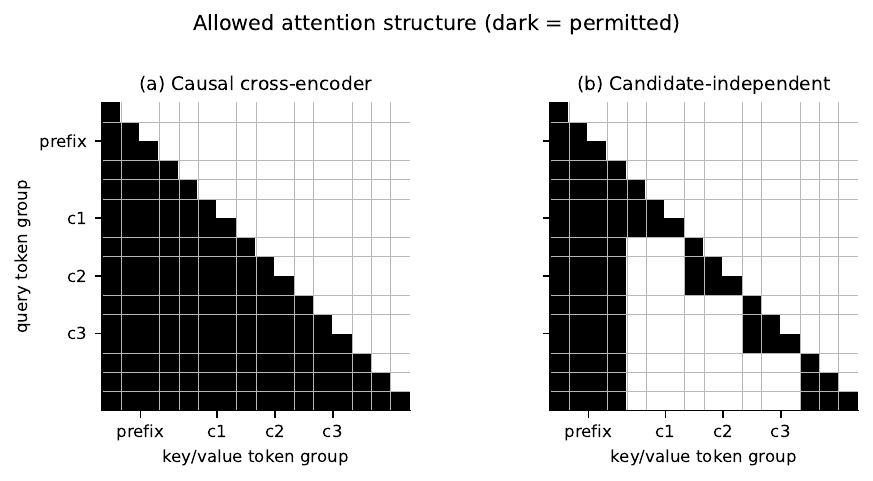}
\caption{Structural attention masks for one prompt with four candidate blocks. Dark cells are permitted attention. Causal attention exposes earlier candidate blocks; CI attention exposes only the shared prefix and the current candidate block.}
\label{fig:masks}
\end{figure*}

Decoder-only transformers compute contextual representations with a causal self-attention mask, in which token $t$ can depend only on tokens at positions no greater than $t$~\cite{vaswani2017attention}. This inductive bias is natural for language modeling and makes autoregressive backbones broadly reusable for decision problems. A common adaptation is to serialize a state, question, and candidate actions into one sequence, then attach a classification or score head to candidate-specific hidden states.

The corresponding causal and candidate-independent attention patterns are shown in Figure~\ref{fig:masks}.

This serialized construction is a cross-encoder: all fields share one forward pass and later tokens can condition on earlier fields. Cross-encoders are expressive because the representation of a candidate may incorporate arbitrary context from the prompt and previously serialized candidates. The same property creates a potential mismatch when the candidate set is semantically unordered. A permutation of candidates changes token positions, causal visibility, and rotary-position phases even when it leaves the underlying decision problem unchanged. Data augmentation over permutations can regularize this shortcut, but it does not remove the information path or impose an exact symmetry.

Set models address this mismatch by requiring permutation invariance or equivariance. Deep Sets show that invariant functions can be represented by shared element encoders followed by a symmetric aggregation~\cite{zaheer2017deep}. In decision ranking, however, a pure pooled representation may discard token-level structure within each candidate. The candidate-independent model studied here keeps causal token processing inside each block and imposes symmetry through the attention graph and local positions rather than through pooling alone.

The closest systems baseline is a dual encoder. Dense retrieval systems encode a query and each candidate independently and compare their embeddings, enabling candidate caching and exact permutation equivariance~\cite{karpukhin2020dense}. Poly-encoder-style designs explore intermediate tradeoffs between independent encodings and full cross-attention~\cite{humeau2020poly}. Our implementation was also inspired by the state/action decomposition and contrastive scoring interface of Contrastive Language Models (CLM)~\cite{clm2026}. The resulting dual baseline is intentionally conservative: it freezes the shared backbone and trains projection heads with an in-batch contrastive objective, providing an invariance and serving reference rather than an optimization-matched comparison with the attention models.

\subsection{Decision formulation}
For a prompt $x$ and candidates $c_1,\ldots,c_K$, the model produces logits and categorical probabilities:
\[
\mathbf{z}=(z_1,\ldots,z_K),\qquad
p_i=\frac{\exp(z_i)}{\sum_{j=1}^{K}\exp(z_j)}.
\]
The supervision is a categorical target over the displayed candidates. We report accuracy, NLL, and multiclass Brier score. Evaluation canonicalizes candidate IDs before comparing predictions across permutations.

\subsection{Architectures}
We compare three scoring architectures. The causal cross-encoder provides full token-level interaction but exposes serialization order; CI removes cross-candidate paths while retaining causal processing within each candidate block; and the dual encoder independently encodes the prefix and candidates, providing exact equivariance and candidate caching. Formal scoring functions, objectives, and the CI mask are given in Section~3.

\section{Candidate-independent architecture}
We first formalize the matched causal and dual scoring functions, then specify the CI mask. In the causal cross-encoder, if $P$ is the shared prefix and $C_{\leq i}$ denotes candidates up to position $i$,
\[
h_i^{\mathrm{causal}}=F_\theta(P,C_{\leq i}),\qquad
z_i=g_\phi(h_i^{\mathrm{causal}}).
\]
The dual encoder instead uses a shared frozen backbone and trainable projection heads:
\[
u=g_P(F_\theta(P)),\qquad
v_i=g_C(F_\theta(C_i)),\qquad
z_i=\frac{u^\top v_i}{\tau},
\]
where the projected vectors are $\ell_2$-normalized. It is trained with
\[
\mathcal{L}_{\mathrm{NCE}}=-\log
\frac{\exp(u^\top v_y/\tau)}
{\sum_{j=1}^{K}\exp(u^\top v_j/\tau)}.
\]
The causal model is expressive for relational decisions but exposes serialization order; the dual model is exactly permutation-equivariant but cannot express token-level candidate interactions. Its frozen backbone, parameterization, and contrastive objective make it a systems and invariance reference rather than an optimization-matched ablation.

Let the serialized input be a shared prefix $P$ followed by candidate blocks $B_1,\ldots,B_K$. For token positions $u$ and $v$, the CI attention mask is defined entrywise as
\[
M_{uv}=\begin{cases}
1, & \text{if }(u,v)\text{ is prefix-causal, prefix-to-candidate,}\\
  & \quad\text{or same-block-causal},\\
0, & \text{if }u\in B_i,\ v\in B_j,\ i\ne j.
\end{cases}
\]

Candidate positions are reset at each block. If $q$ is the serialized position and $s_i$ is the first position of block $i$, the logical position is $p=q-s_i$ for all tokens in $B_i$. Thus the same candidate text receives the same rotary-position sequence regardless of displayed index, while the prefix retains its original positions.

The serialized template in Figure~\ref{fig:tokens} illustrates how special tokens delimit the shared context, candidate blocks, and score location. Each candidate produces a local representation $H_i$ from its final candidate token or designated score token. A shared scoring head produces $z_i=g(H_i)$, and
\[
p_i=\frac{\exp(z_i)}{\sum_{j=1}^{K}\exp(z_j)}.
\]
Representations are independent, but the selected action still depends on all candidates through the softmax denominator.

For a candidate-block permutation $\pi$, the intended symmetry is
\[
f(T_\pi x)=\Pi f(x),
\]
where $\Pi$ reorders output coordinates. Causal attention has no structural guarantee of this identity; a dual encoder satisfies it exactly by independent encoding. CI preserves token-level causal context while enforcing the same block-level information-flow constraint.

\begin{figure}[t]
\centering
\fcolorbox{gray}{gray!10}{%
\parbox{0.96\linewidth}{%
\ttfamily\footnotesize
\textless BOS\textgreater\ \textless STATE\textgreater robot at (3, 4) \textless/STATE\textgreater\\
\textless QUESTION\textgreater choose the next action \textless/QUESTION\textgreater\\
\textless CANDIDATE id="1"\textgreater move north \textless/CANDIDATE\textgreater\\
\textless CANDIDATE id="2"\textgreater move east \textless/CANDIDATE\textgreater\\
\textless CANDIDATE id="3"\textgreater wait \textless/CANDIDATE\textgreater\\
\textless SCORE\textgreater
}}
\caption{Example text template with special tokens. Candidate delimiters define blocks for masking and logical position resets; the score token denotes the final candidate competition.}
\label{fig:tokens}
\end{figure}

\section{Experimental Setup}
\strong{Data and matched comparisons.} The primary comparison uses the Open-Jev typed-decision dataset~\cite{openjev} with identical train, validation, test, and out-of-distribution (OOD) partitions for every architecture. Examples retain their source and group identifiers throughout preprocessing, so models are evaluated on the same decisions rather than on architecture-specific resamples. The held-out test sets contain 5,000 examples for Gemma 3 1B and Qwen3 1.7B and 10,208 examples for Qwen3 4B; the shared OOD set contains 15,539 examples. The Gemma 3 1B mechanism ablation uses the full Open-Jev training split and contrasts causal attention, full CI, isolation-only, position-reset-only, and CI without permutation augmentation. Dual-encoder runs are reported as historical systems baselines because their frozen-backbone, contrastive training recipe is not matched in optimization to the attention models.

\strong{Training and checkpoint selection.} All matched models are trained for two epochs with the same effective global batch size of 64, BF16 backbone computation, FP32 classification heads and loss evaluation, and the same optimizer schedule. We select the checkpoint with the lowest validation NLL rather than the final checkpoint, avoiding a comparison that depends on different training durations or arbitrary stopping points. This protocol keeps capacity and data exposure fixed while allowing each architecture to use its own valid forward pass and memory-efficient batching.

\strong{Evaluation and uncertainty.} Each held-out example is scored in its original candidate order and under five fixed candidate permutations. Predictions are mapped back to canonical candidate IDs before metrics are computed. We report accuracy, negative log-likelihood (NLL), multiclass Brier score, calibration error, and confidence margins. To quantify order sensitivity, we additionally report prediction flip rate, mean Jensen--Shannon divergence between permutation distributions, mean absolute probability change, and worst-permutation accuracy. Since multiple decisions can belong to the same latent group, paired uncertainty uses 5,000 group-level bootstrap resamples (seed 2026), rather than treating rows as independent. This preserves the pairing between architectures and prevents repeated variants of one underlying task from being counted as unrelated evidence.

\section{Results}
\subsection{Matched backbone comparisons}
Table~\ref{tab:main} summarizes the completed matched runs. CI improves test accuracy for all three backbones and remains within 0.72 points of causal on OOD accuracy. Its defining advantage is order robustness: test flip rates fall from 8.3\% to 0.1\% for Gemma 3 1B, 18.2\% to 1.2\% for Qwen3 1.7B, and 7.2\% to 0.7\% for Qwen3 4B.

\begin{table*}[t]
\centering
\caption{Matched causal and CI results. Lower NLL/Brier/flip is better. Bold marks the better value within each backbone and split.}
\label{tab:main}
\small
\resizebox{0.90\textwidth}{!}{%
\begin{tabular}{llrrrrrrr}
\specialrule{.8pt}{0pt}{1pt}\specialrule{.35pt}{0pt}{1pt}
Backbone & Model & \multicolumn{3}{c}{Test} & \multicolumn{3}{c}{OOD} \\
 & & Acc. & NLL & Brier & Acc. & NLL & Brier & Flip\% \\
\midrule
Gemma 3 1B & Causal & 84.38 & .335 & .178 & 80.44 & \best{.640} & \best{.291} & 8.30 \\
 & CI & \best{84.66} & \best{.327} & \best{.171} & \best{80.71} & .723 & .301 & \best{.11} \\
Qwen3 1.7B & Causal & 81.84 & .388 & .209 & \best{75.09} & \best{.706} & \best{.358} & 18.16 \\
 & CI & \best{84.96} & \best{.348} & \best{.185} & 74.37 & .735 & .375 & \best{1.20} \\
Qwen3 4B & Causal & 91.05 & .208 & .095 & 83.43 & \best{.738} & .258 & 7.19 \\
 & CI & \best{92.22} & \best{.180} & \best{.080} & \best{83.83} & .777 & \best{.256} & \best{.67} \\
\specialrule{.35pt}{1pt}{0pt}\specialrule{.8pt}{1pt}{0pt}
\end{tabular}
}
\end{table*}

\begin{figure}[t]
\centering
\includegraphics[width=\linewidth]{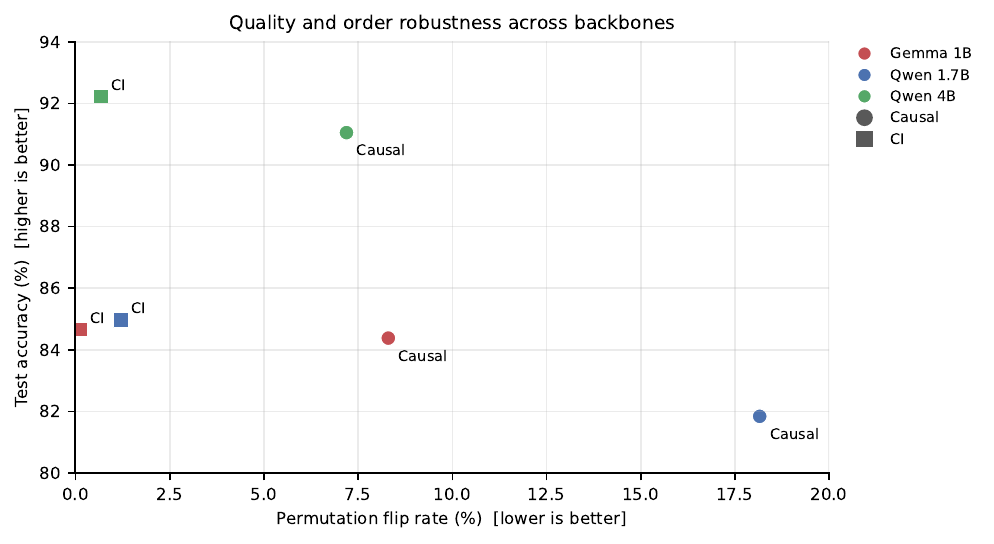}
\caption{Test accuracy versus permutation flip rate. Colors denote backbones; circles are causal and squares are CI. CI consistently reduces flips without a systematic accuracy penalty.}
\label{fig:tradeoff}
\end{figure}

Figure~\ref{fig:tradeoff} shows the central quality/robustness pattern. CI moves every point toward the lower-left in flip rate while maintaining accuracy. On Qwen3 1.7B, the test improvement is large (84.96 versus 81.84\%); on Gemma 3 1B and Qwen3 4B, the quality difference is small. OOD differences are not uniformly directional, so order robustness should not be conflated with distribution-shift robustness.

\subsection{Paired inference and heterogeneity}
For Qwen3 1.7B, group bootstrap favors CI on all three test metrics: accuracy difference $+3.12$ points (95\% CI $[+0.82,+5.63]$), NLL difference $-0.039$ ($[-.072,-.010]$), and Brier difference $-0.024$ ($[-.045,-.004]$). All OOD intervals cross zero. The Gemma 3 1B rerun is similarly near-tied in accuracy; causal improves OOD NLL by $0.082$ relative to CI. Qwen3 4B test quality favors CI by 1.18 points of accuracy and 0.028 NLL, while OOD accuracy is nearly tied.

The gains are heterogeneous. Causal is stronger on some binary and large-candidate slices, whereas CI is particularly strong on Qwen3 1.7B three- and five-candidate decisions and on VizDoom. We therefore treat candidate-count and source slices as descriptive rather than pooled evidence.

\subsection{Mechanism ablation}
\begin{figure}[t]
\centering
\includegraphics[width=\linewidth]{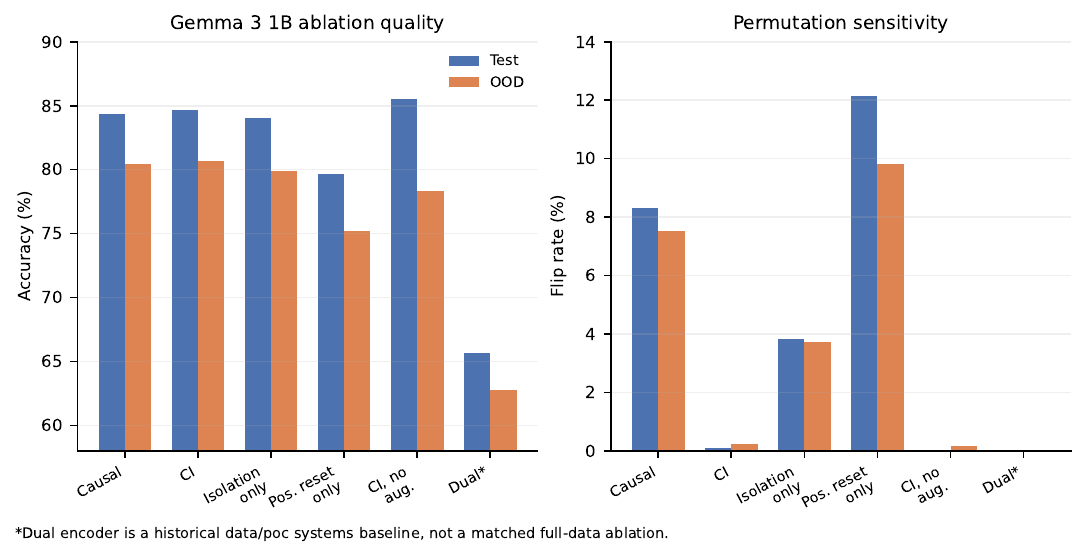}
\caption{Gemma 3 1B full-data mechanism ablation. Isolation is the primary robustness mechanism; position resetting alone is not sufficient. The dual row is a historical systems baseline.}
\label{fig:ablations}
\end{figure}

The Gemma 3 1B ablation (Figure~\ref{fig:ablations}) isolates the two structural ingredients. Isolation-only reduces test flips from 8.30\% to 3.82\% and OOD flips from 7.54\% to 3.71\%, but does not remove sensitivity because global candidate positions remain. Position-reset-only is worse than causal (12.16\% test flips and 79.64\% accuracy), showing that shared positions without isolation are insufficient. Removing training-time permutation augmentation preserves near-invariance (0.04\% test flips) but changes OOD quality, indicating that the architectural mask supplies the main invariance while augmentation affects optimization and coverage. The dual encoder is exactly invariant, but its historical proof-of-concept run reaches only 65.68\% test accuracy and is not an optimization-matched ablation.

\section{Qwen3-4B 250K research artifact}
This artifact extends the architecture comparison from a single curated source to a heterogeneous public mixture. It is a candidate-independent Qwen3-4B model trained on 250,000 decision examples and is intended as a reproducibility and coverage study rather than as a new causal-versus-CI comparison. The released checkpoint and tokenizer are available at the accompanying \href{https://huggingface.co/Guy-Amit/qwen3-4b-ci-decision-4096-poc}{\textcolor{blue}{Hugging Face artifact page}}. The longer 4096-token context is part of the artifact definition: the matched Qwen3 4B Open-Jev runs use a 1024-token limit, so their scores are not directly interchangeable.

The mixture combines six sources with deliberately different decision formats and risk profiles: Open-Jev contributes 79,116 examples~\cite{openjev}, xLAM 57,000~\cite{xlam}, Aegis 2.0 30,007~\cite{aegis2}, ToolACE 10,700~\cite{toolace}, repository prompt injection 5,387~\cite{promptrepo}, and WildGuardTrain 67,790~\cite{wildguard}. The validation set is stratified by source, with 3,000 examples per subset for 18,000 total. Group identifiers and source labels are retained so that source-level estimates remain paired and auditable rather than being reconstructed from aggregate counts.

The model uses \texttt{Qwen/Qwen3-4B} with a 4096-token context and is trained for 4,000 optimizer steps. Backbone and head learning rates are $5\times10^{-6}$ and $5\times10^{-5}$, respectively; optimization uses BF16 computation, 8-bit AdamW, weight decay 0.01, a 0.05 warmup ratio, gradient clipping at 1.0, and seed 42. Candidate permutation augmentation is enabled. A per-device batch of 32 with two accumulation steps on one GPU gives an effective batch of 64. Validation is evaluated periodically, and model selection uses validation NLL.

The source composition and balanced validation results for this artifact are reported in Table~\ref{tab:artifact}.

\begin{table}[t]
\centering
\caption{Qwen3-4B 250K artifact coverage. Validation is balanced across the six training sources; aggregate OOD results are reported below.}
\label{tab:artifact}
\scriptsize
\resizebox{\columnwidth}{!}{%
\begin{tabular}{lrrrll}
\specialrule{.8pt}{0pt}{1pt}\specialrule{.35pt}{0pt}{1pt}
Subset & Train & Val. & Val. Acc. & Val. NLL & Val. Flip\\
Open-Jev & 79,116 & 3,000 & 93.10\% & 0.1721 & 0.20\%\\
xLAM & 57,000 & 3,000 & 100.00\% & 0.0000 & 0.00\%\\
Aegis 2.0 & 30,007 & 3,000 & 97.50\% & 0.0628 & 0.03\%\\
ToolACE & 10,700 & 3,000 & 100.00\% & 0.0002 & 0.00\%\\
Repo injection & 5,387 & 3,000 & 85.03\% & 0.3239 & 0.83\%\\
WildGuardTrain & 67,790 & 3,000 & 98.00\% & 0.0492 & 0.03\%\\
\specialrule{.35pt}{1pt}{0pt}\specialrule{.8pt}{1pt}{0pt}
\end{tabular}}
\end{table}

\begin{figure}[t]
\centering
\includegraphics[width=\linewidth]{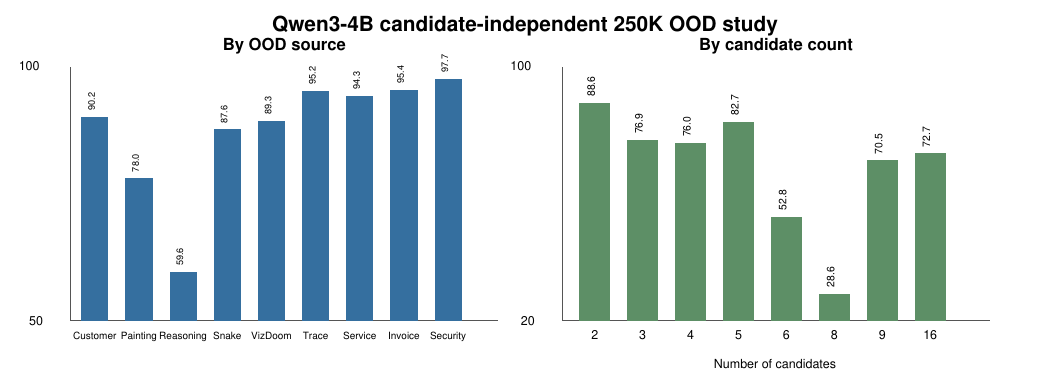}
\caption{Qwen3-4B CI 250K validation and OOD results. Validation uses 3,000 examples from each training source; OOD accuracy is shown by source and candidate count on a separate benchmark mixture.}
\label{fig:scale}
\end{figure}

On the balanced validation split, the artifact reaches 95.61\% accuracy, 0.1014 NLL, 0.0595 multiclass Brier score, and 0.0085 top-label ECE. The validation audit records a 0.183\% permutation flip rate, mean Jensen--Shannon divergence of $1.24\times10^{-5}$, and maximum per-example probability drift of 0.0776. On OOD, it reaches 83.84\% accuracy, 0.6762 NLL, 0.2519 Brier score, and 0.0044 permutation flip rate, with a mean Jensen--Shannon divergence of $4.47\times10^{-5}$. OOD accuracy is 88.61\% for binary decisions, 76.86\% for three candidates, 82.72\% for five candidates, and 72.69\% for sixteen candidates; the eight-candidate slice is the most difficult at 28.57\%. The aggregate and slice results are visualized in Figure~\ref{fig:scale}. Together, these results indicate transfer across decision formats while preserving order robustness under distribution shift.

\section{Related methods and discussion}
The causal baseline follows the standard decoder-only transformer formulation~\cite{vaswani2017attention}. Its candidate interaction is analogous to a cross-encoder: all text is jointly represented, which is often favorable when pairwise or listwise interactions matter. The cost is that sequence order can become a shortcut. Permutation augmentation is a common practical mitigation, but it does not guarantee invariance when the mask and positional encoding still expose order.

CI instead builds a partial-information graph into attention. This is related to permutation-equivariant set processing~\cite{zaheer2017deep} and masked transformers, but differs from pooling-based set encoders: candidates retain token-level causal structure and compete only at the decision head. The dual encoder follows the independent-encoding family used in dense retrieval~\cite{karpukhin2020dense}; it offers exact equivariance and candidate caching, but independent representations can discard interactions needed for difficult listwise decisions. Our results make this tradeoff concrete: the dual baseline is robust but weak on nonbinary and high-cardinality choices.

\strong{Why not simply sort or average?} Sorting candidates requires a task-specific canonical order and can discard uncertainty about ties. Averaging scores over permutations reduces sensitivity at inference but multiplies compute and does not prevent the model from learning order-dependent features. CI imposes the information-flow constraint directly and requires only one forward pass at serving time.

The quality tradeoff is architectural rather than universal. Causal interaction can help when the correct action depends on explicit candidate comparisons, whereas CI forces those comparisons through the final competition layer. These distinctions explain why the matched results are heterogeneous across sources and candidate counts; they are not evidence that one architecture dominates in every regime.

Our claim is deliberately narrow: candidate-independent masking transfers order robustness across backbones at comparable decision quality. It does not establish computational sparsity, universal quality gains, or the best possible objective-matched dual encoder. Those questions require optimized block-sparse kernels, additional seeds, and a separately tuned dual-encoder training recipe.

\section{Limitations and conclusion}
Our study is intentionally focused on the architectural source of candidate-order sensitivity. The main comparisons use one training seed per condition, and the CI mask is evaluated with dense transformer kernels; additional seeds and block-sparse implementations would help quantify variance and serving efficiency. The dual encoder is included as a useful reference point, but its objective and parameterization are not fully matched to the causal and CI models. Finally, the 250K case study examines the CI model on balanced validation and OOD data, rather than providing a matched causal data-scale comparison. These choices delimit the scope of the evidence and suggest direct directions for follow-up work.

Taken together, the results provide a consistent positive signal. Across Gemma 3 1B, Qwen3 1.7B, and Qwen3 4B, candidate-independent block-causal attention substantially reduces permutation sensitivity while preserving competitive decision quality. The mechanism ablation indicates that candidate isolation is the main contributor, with position resetting completing the intended symmetry. Causal attention remains useful when explicit candidate interaction is important, while dual encoders remain attractive for exact invariance and caching. For System 1 decision models inside generative systems, CI makes candidate ordering a controlled architectural property rather than an accidental feature of serialization.

\balance

\end{document}